\documentclass{article}

\usepackage{arxiv}

\usepackage[utf8]{inputenc} 
\usepackage[T1]{fontenc}    
\usepackage{hyperref}       
\usepackage{url}            
\usepackage{booktabs}       
\usepackage{amsmath,amssymb,amsfonts}      
\usepackage{algorithmic}
\usepackage{textcomp}
\usepackage{bbm}
\usepackage{booktabs}

\usepackage{nicefrac}       
\usepackage{microtype}      
\usepackage{lipsum}
\usepackage{fancyhdr}       
\usepackage{graphicx}       
\usepackage{multirow}
\graphicspath{{media/}}     

\title{NRTR: Neuron Reconstruction with Transformer from 3D Optical Microscopy Images
}

\author{
  Yijun Wang$^1$, Rui Lang$^2$  \\
  School of Informatics \\
  Xiamen University \\
  Xiamen, China\\
  \texttt{\{wangyijun, langrui\}@stu.xmu.edu.cn} \\
   \And
  Rui Li$^3$ \\
  National Engineering Laboratory for Educational Big Data \\
  Central China Normal University \\
  Wuhan, China\\
  \texttt{leerui@ccnu.edu.cn} \\
    \And
  Junsong Zhang$^{4*}$ \\
  School of Informatics \\
  Xiamen University \\
  Xiamen, China\\
  \texttt{zhangjs@xmu.edu.cn} \\
}

\begin{document}
\maketitle

\begin{abstract}
The neuron reconstruction from raw Optical Microscopy (OM) image stacks is the basis of neuroscience. Manual annotation and semi-automatic neuron tracing algorithms are time-consuming and inefficient. Existing deep learning neuron reconstruction methods, although demonstrating exemplary performance, greatly demand complex rule-based components. Therefore, a crucial challenge is designing an end-to-end neuron reconstruction method that makes the overall framework simpler and model training easier. We propose a Neuron Reconstruction Transformer (NRTR) that, discarding the complex rule-based components, views neuron reconstruction as a direct set-prediction problem. To the best of our knowledge, NRTR is the first image-to-set deep learning model for end-to-end neuron reconstruction. In experiments using the BigNeuron and VISoR-40 datasets, NRTR achieves excellent neuron reconstruction results for comprehensive benchmarks and outperforms competitive baselines. Results of extensive experiments indicate that NRTR is effective at showing that neuron reconstruction is viewed as a set-prediction problem, which makes end-to-end model training available.
\end{abstract}

\keywords{Neuron reconstruction \and Set prediction problem \and SWC reconstruction file \and Transformer}

\section{Introduction}
\label{sec:introduction}
Neuron reconstruction aims to extract neuron topological information from raw neuron image stacks. For neuroscience research, high-quality neuron reconstruction results are fundamental and essential.

With the application of deep learning methods to Computer Vision (CV) fields, it has been proved that deep learning models significantly affect different tasks. Many of the proposed neuron reconstruction models, based on deep learning, have been proposed and achieved significant results. In various CV tasks, Convolution Neural Network (CNN) has dominated for many years. As far as we know, Li \textit{et al.} \cite{li2017deep} proposed the first 3D CNN-based U-Net \cite{ronneberger2015u} for neuron reconstruction. They tend to cast neuron reconstruction as a medical image segmentation problem. Following their work, many researchers \cite{huang2020weakly, chen2021weakly, yang2021structure} proposed various versions of U-Net to produce neuron segmentation results.

However, some problems remain to be solved. These U-Net-based neuron reconstruction methods indirectly address the neuron reconstruction problem by combining image segmentation methods and rule-based neuron tracing algorithms, including APP2 \cite{xiao2013app2}, MOST \cite{ming2013rapid}, MEIT \cite{wang2018memory}, FMST \cite{yang2019fmst}, Rivulet \cite{liu2016rivulet}, as post-processing. The performance of these methods is influenced significantly by those rule-based neuron tracing algorithms; whereas, deep learning models seem to be applied only for image enhancement or image denoising. The difficulties caused by these methods involve a complex pipeline and the impossibility of end-to-end model training. Moreover, as the core mechanism of CNN, the convolution operation is somewhat limited in capturing global features. We used to propose a probabilistic method named G-Cut that automatically segments individual neurons from a dense neuron cluster \cite{li2019precise}. Like the deep learning methods mentioned above, G-Cut is based on rule-based neuron tracing algorithms to improve the quality of neuron reconstructions.

In recent years, CV has witnessed the rapid development of deep learning model architectures, wherein a Transformer is introduced to CV. Different from convolution operations, the self-attention mechanism of the Transformer is adept at extracting global image information. Besides, the Transformer is a typical sequence-to-sequence model. With some modification, a variant of the Transformer can embed a neuron image stack to a point set that represents the neuronal structure.

\begin{figure*}[ht]
\centerline{\includegraphics[width=\textwidth]{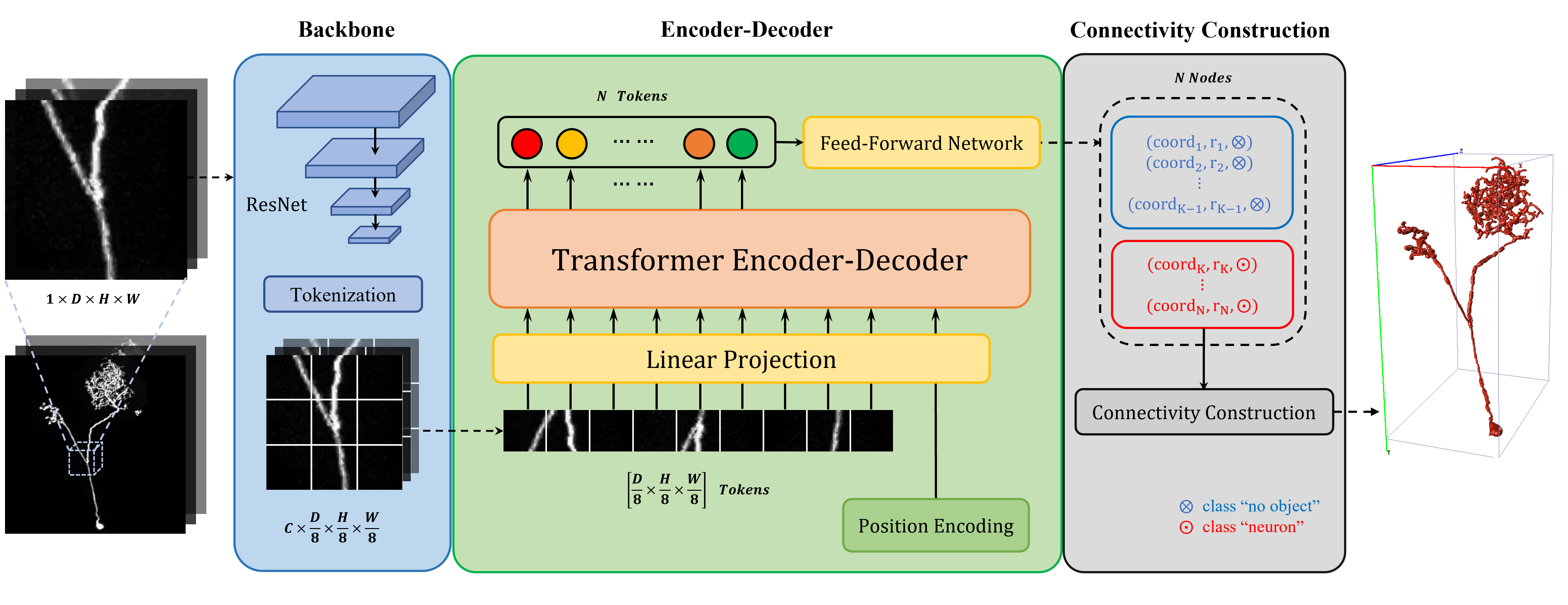}}
\caption{
The pipeline of our proposed method. Due to the massive size of the raw OM neuron image stacks and the limitation of memory size, the whole image stacks are cropped into several fixed-size image blocks. NRTR extracts the high-dimensional feature map from each image block through the CNNs backbone. Then the feature map is split into patches, and the sequence of linear projections of these patches is provided as an input to a Transformer encoder-decoder. Finally, the prediction head directly generates neuron reconstruction results (point sets). 
In the training stage, bipartite matching assigns prediction points and ground truth points. To align two set sizes, class "no-object" ($\varnothing$) is introduced into the predicted class. The point is either "neuron" (red) or "no-object" (blue).}
\label{Fig-1}
\end{figure*}
 
To streamline the overall neuron reconstruction pipeline, we formulate neuron reconstruction as a direct set-prediction problem. Using fully these two properties of a Transformer, we propose the first image-to-set model for end-to-end neuron reconstruction (NRTR), which mitigates the two issues mentioned above. Unlike previous U-Net-based deep learning methods, NRTR generates a point set from a neuron image stack. As shown in Fig. \ref{Fig-1}, NRTR is a hybrid model combining the CNN backbone and Transformer encoder-decoder. The overall pipeline does not involve any complex rule-based components. Extensive experiments on various datasets show that our method achieves significant results on comprehensive benchmarks.

Our major contributions are as follows:
\begin{enumerate}
\item The key contribution of our work is the formulation of neuron reconstruction as a set-prediction problem by minor modification of Transformer architecture. We construct the connection relationship among points through a connectivity construction module. The new model is conceptually simple and does not require any extra hand-designed components.
\item Proposed for neuron reconstruction is a novel image-to-set model, which achieves excellent results with comprehensive benchmarks. Compared with previous U- Net-based models, NRTR can be trained end-to-end for neuron reconstruction, indicating that we are one step closer to fully-automatic neuron reconstruction.
\end{enumerate}

\section{Related Work}
\label{sec:related work}
Our work is based on prior work in the following four domains: neuron reconstruction methods, standard SWC reconstruction file, set-prediction problem, and Transformer architecture.

\subsection{Neuron Reconstruction}
Characterizing 3D neuron morphology is the basis of neuroscience research. For neuron reconstruction, many researchers recently have proposed various rule-based neuron tracing algorithms, or deep learning methods.

Many efforts have been devoted to developing rule-based algorithms for OM image stacks. Some algorithms \cite{xiao2013app2,ming2013rapid,wang2018memory,yang2019fmst,liu2016rivulet} tend to filter out background noises from OM image stacks through binarization and generate corresponding neuron reconstructions. Generally inspired by graph theory and neuron morphology theory, these methods are highly interpretable. However, most weak neuron signals in raw image stacks are filtered as noises because of binarization \cite{chen2021weakly}. These methods focus mainly on a single neuron and do not perform well with dense neuron images. Thus, these rule-based neuron tracing algorithms remain a bottleneck.

Researchers have proposed deep learning methods to mitigate weakness by replacing binarization with image segmentation models. These approaches predict a probability map through deep segmentation networks and feed it to one of the rule-based neuron tracing algorithms mentioned above.

Many researchers have proposed recently various deep learning models to extract neuron signals and remove background noise. Li \textit{et al.} \cite{li2017deep} designed 3D U-Net to enhance raw OM image stacks for more high-quality neuron reconstructions. Liu \textit{et al.}  \cite{liu2018improved} proposed an improved V-Net to enhance the neuronal structure and reduce the influence of noises. Bo \textit{et al.} \cite{yang2021structure} presented a 3D neuron segmentation network, SGSNet, which contains a shared encoding path and two decoding paths. To solve the problem of insufficient annotated neuron image stacks, Chen \textit{et al.} \cite{chen2021weakly} and Huang \textit{et al.} \cite{huang2020weakly} proposed a weakly supervised learning method (MP-NRGAN) and a progressive learning method (PLNPR).

Other researchers have proposed probabilistic methods to address the neuron reconstruction task. Li \textit{et al.} \cite{li2019precise} proposed using a probabilistic method, G-Cut, that constructs more accurate connections among somas and improves neuron reconstruction results. Athey \textit{et al.} \cite{athey2022hidden} presented a novel method, ViterBrain, which employs a hidden Markov state process that encodes neuron geometry with a random field appearance model.

We believe that the hybrid frameworks do not take full advantage of deep learning methods, and image segmentation models are improved binarization. Rule-based neuron tracing algorithms play a more critical role in the overall framework. Therefore, for neuron reconstruction, we propose a novel deep learning framework which completely discards rule-based neuron tracing algorithms and achieves end-to-end model training and inference.
 
\subsection{SWC Reconstruction File}
Researchers, to detail complex neuronal structures, manually trace the neuron morphology using optical microscopy and digitally reconstruct neuronal organization and arrangement through microscopy images.

The SWC neuron morphology format is widely used to store these digital neuron reconstructions \cite{o2020module}. One location in this volume is defined as the root for each image stack. Digital reconstruction consists of one or more trees starting at the roots. Each tree is represented as a set of interconnected tapering points or compartments. A root is assigned a virtual parent with identity -1.

Each point is represented in the reconstruction file by a line of seven fields: a numeric identity, an optional tag (including soma, axon, and dendrite), the 3D center coordinates, and the radius and the identity of its parent point \cite{cannon1998line}. Thus, the SWC reconstruction file can be viewed as a point set. Compared with raw OM images, the SWC files consume little storage space and clearly express the connectivity among points. Note that the SWC neuron representation, widely adopted for manual reconstruction from optical microscopy neuron image stacks, differs from the surface rendering, resulting from membrane contours commonly traced in electron microscopy neuron image stacks.

\subsection{Set Prediction}
It is expected that our proposed model will generate a point set consistent with the SWC file format. Each point consists of 3D center coordinates and radius. Then, the connectivity among points is built through a rule-based algorithm that embeds a point set to a tree or more trees. In this way, neuron reconstruction is formulated as a set-prediction problem.

Previous deep learning models, although designed to solve the set-prediction problem, have not been introduced into neuron reconstruction. The difficulty with the set-prediction problem is how to avoid a set filled with similar or repeated elements. 

Some prior work resorted to an autoregressive model to generate the set. An autoregressive model generates each element one by one, conditioned on the elements previously generated, the weakness of which is the generation is time-intensive. Thus, various non-autoregressive models are proposed to solve the set-prediction problem more efficiently. DETR \cite{carion2020end}, which uses a specific set-based loss, is a typical non-autoregressive model for the set-prediction problem. The Hungarian algorithm is a solution to searching for a bipartite matching between ground truth elements and prediction elements. Based on bipartite matching, the set-based loss is denoted as the sum of the loss value between each element in the prediction set and the corresponding one in the ground truth set. The bipartite matching and the set-based loss are used to enlarge the distance among the embedded vector of elements. Hence, a set-prediction model generates a set with various elements.

\subsection{Vision Transformer}
Since it was proposed, the Transformer, which shows impressive progress in machine translation, document summarization, text classification, and more, has become the standard for various natural language processing tasks \cite{dosovitskiy2020image}. The transformer model and self-attention mechanism have revolutionized machine translation and natural language processing. Compared with prior work, the Transformer is a model architecture that, instead of eschewing recurrence, relies entirely on an attention mechanism to draw global dependencies between input and output \cite{NIPS2017_3f5ee243}.

Our proposed method NRTR is greatly inspired by DETR \cite{carion2020end}, proposed by Carion \textit{et al.}, and Vision Transformer (ViT) \cite{dosovitskiy2020image}, proposed by Dosovitskiy \textit{et al.}. ViT demonstrates that, by the specific tokenization of an image, a Transformer-based model is an option as a visual backbone in CV tasks. An image is split into several fixed-size tokens (also called patches). Similar to ViT, DETR partitions the image into several tokens. The key innovation of DETR is that object detection is regarded as a direct set-prediction problem. DETR is a hybrid model, which combines the CNN backbone and the Transformer encoder-decoder. Given learned object queries, DETR infers the relationships and differences among objects to predict a fixed-size set of bounding boxes.

 Inspired by DETR and relative work, our method regards neuron reconstruction as a set-prediction problem. Thus, without any rule-based neuron tracing algorithms, NRTR can be trained end-to-end. Through the connectivity construction, neuron morphological characteristics, predicted by NRTR, are in the SWC format. In addition, topological characteristics of neurons are more clearly represented by a tree, rather than by a probability map.

\begin{figure*}[ht]
\centerline{\includegraphics[width=\textwidth]{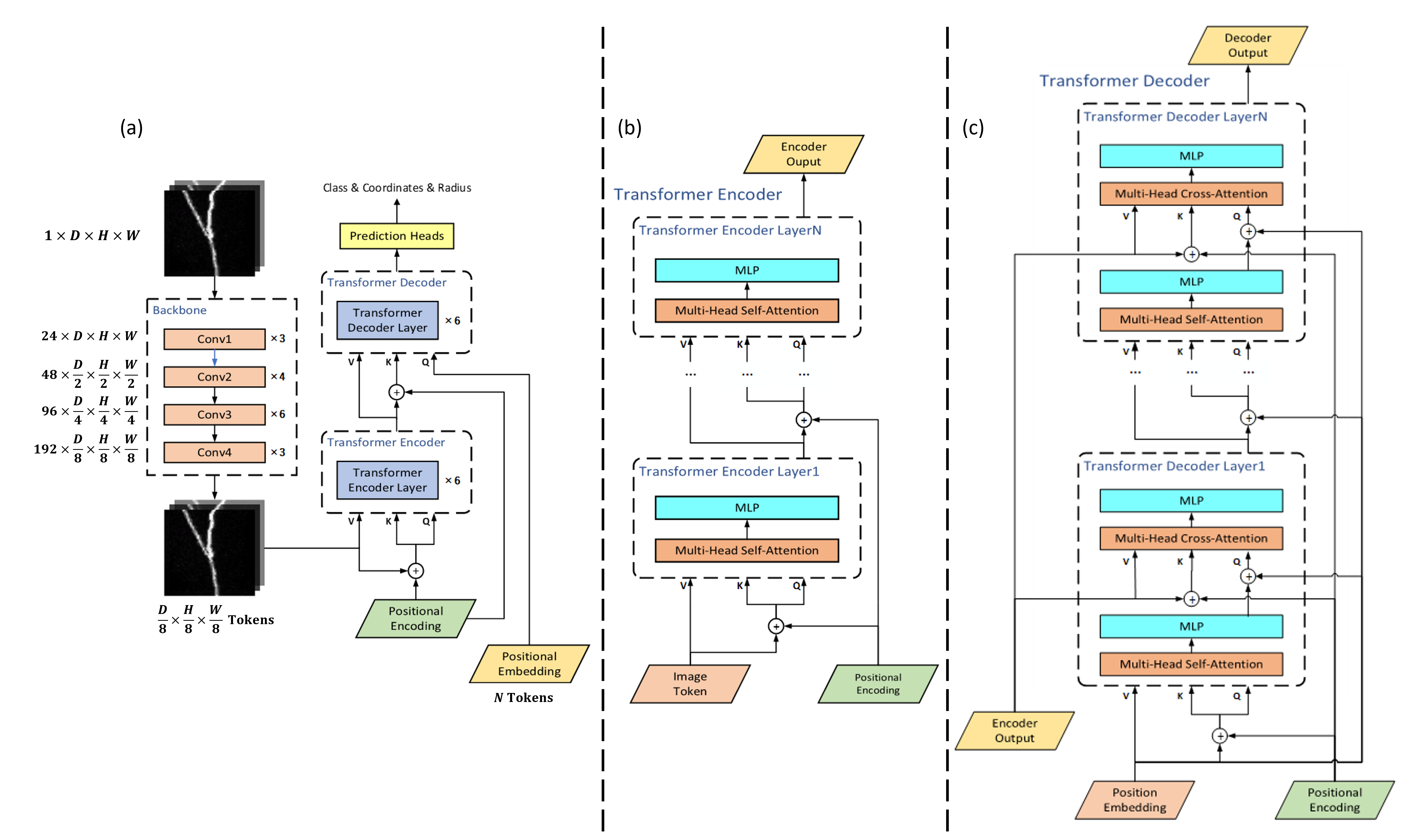}}
\caption{(a) The overall architecture of NRTR. (b) The Transformer encoder. (c) The Transformer decoder. We employ the standard ResNet with FPN to extract multi-scale feature maps. The Transformer encoder takes tokenized feature maps as input and produces encoded tokens. The Transformer decoder takes encoded tokens and a fixed number of positional embeddings as input and produces a fixed number of output embeddings. Finally, the prediction head generates one point with 3D center coordinates, radius, and class from each output embedding.}
\label{Fig-3}
\end{figure*}

\begin{figure}[ht]
\centering
\centerline{\includegraphics[width=\columnwidth]{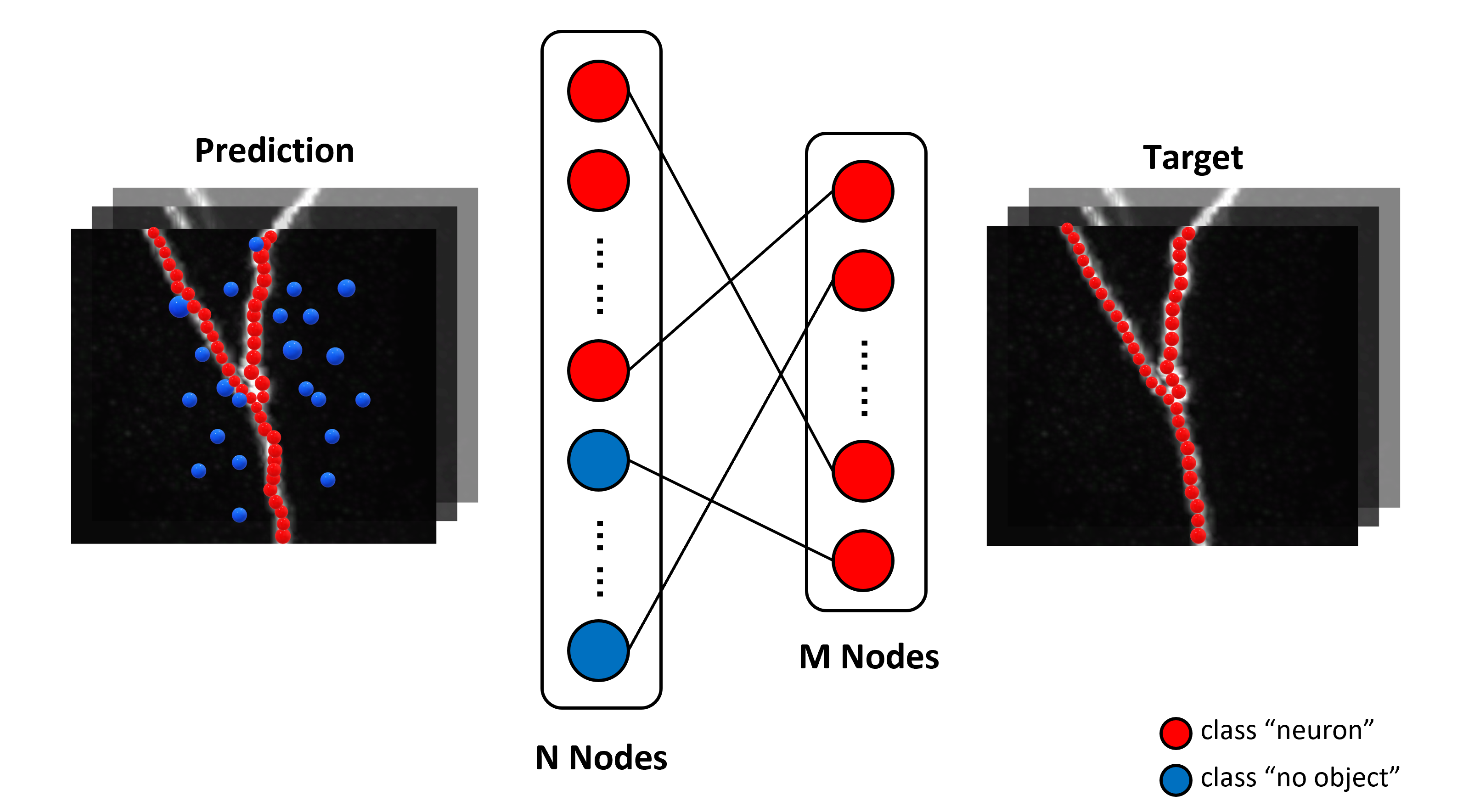}}
\caption{NRTR generates a fixed-size set of N points, while the ground truth set contains M points. Before assessing the loss, the Hungarian algorithm is used for a minimum weight bipartite matching between the prediction and ground truth sets. M best-matched points of N prediction points are calculated into loss, which indicates that N is larger than M.}
\label{Fig-4}
\end{figure}

\section{Method}
\label{sec:method}
For most previous deep learning methods \cite{huang2020weakly, chen2021weakly, yang2021structure}, where neuron reconstruction is regarded as a medical image segmentation problem, various U-Nets were proposed that are common in image segmentation. However, we believe that a method will solve the problem more conveniently and achieve better neuron reconstruction results if we view neuron reconstruction as a direct set-prediction problem.

Thus, motivated for neuron reconstruction, we propose the novel method, NRTR. NRTR accomplishes neuron reconstruction end-to-end; whereas, neuron image segmentation methods (including MP-NRGAN \cite{chen2021weakly}, SGSNet \cite{yang2021structure}, and PLNPR \cite{zhao2020neuronal}) take rule-based algorithms (including APP2 \cite{xiao2013app2} and MOST \cite{ming2013rapid}) as the neuron tracing unit. To mitigate the issue, our model prefers to infer neuron morphology from OM neuron image stacks by directly generating a point set, similar to a SWC file. This way, complex rule-based components can be removed and the entire neuron reconstruction pipeline simplified.

NRTR is expected to produce a point set for each image block cropped from the raw OM image stack $x=\{x_i\}_{i=1}^{1\times W\times H \times D}$, where $W, H, D$ are the width, height, and depth of the image block.
For every block, NRTR predicts the point set, denoted as 
\begin{equation}
y=\{(a_j, b_j, c_j, r_j, cls_j)\}_{j=1}^N\label{eq1}
\end{equation}
where $N$ is a predefined constant; $(a_j, b_j, c_j)$ and $r_j$ are the normalized center coordinates and radius of the point, respectively; $cls_j \in [0, 1]$ is the probability of the point belonging to neurons.

The overall NRTR architecture consists of three main components: backbone, Transformer, and connectivity construction. The architecture of our deep learning method, divided into three parts, is described in detail in Fig.\ref{Fig-3}.

\subsection{Backbone}
Given an initial image stack, $ x_{img}\in \mathbb{R}^{1\times H_o \times W_o \times D_o}$, the standard ResNet \cite{he2016deep} with FPN \cite{lin2017feature} is employed as the backbone to extract the corresponding lower-resolution feature map, $ x_f\in \mathbb{R}^{C\times H \times W \times D}$. This way, because our model recognizes the initial image at vastly different scales, a multi-scale feature map is incorporated into the neuron reconstruction.

The Transformer, introduced below, is a sequence process unit. Before being fed to the Transformer, features are sequentially extracted by the ResNet to create the feature map, $x_f$, which is partitioned into a sequence, $s_o$.
\begin{equation}
s_o =(s_1, s_2,\dots ,s_n) , s_i \in \mathbb{R}^C
\end{equation}
where $n$ is the sequence length; and $C$ is the channel dimension. Because the multi-head self-attention mechanism is permutation-invariant, to take spatial information into account, positional encoding is added to the feature sequence \cite{parmar2018image}:
\begin{equation}
s =(s_1+p_1, s_2+p_2,\dots ,s_n+p_n) , s_i\in \mathbb{R}^{C}, p_i\in \mathbb{R}^{C}
\end{equation}
where the feature map, $s_i$, is the aggregate of the sequential features, and $p_i$ is the corresponding positional encoding. Typically, the hyper-parameters, denoted above, are $C = 192$ and $H, W, D = \frac{H_0}{8} , \frac{W_0}{8}, \frac{D_0}{8}$, respectively.

\subsection{Transformer}
The Transformer module has two major components: Transformer encoder and Transformer decoder. Given the sequence input $s$ extracted by a ResNet backbone, the Transformer encoder transforms the input sequence $(s_1, s_2\dots ,s_n)$ into a state $z$. Then, the Transformer decoder maps the encoded state, $z$, to a fixed-size sequence, $y=(y_1, y_2,\dots,y_m)$ as output. In contrast to the standard Transformer \cite{vaswani2017attention}, the entire sequence ($N$ points) is generated simultaneously, which reduces the time cost of model training and inference.

The Transformer encoder comprises $K$ encoder layers, where $K$ is the hyperparameter. Each encoder layer comprises a multi-head self-attention module and a Multi-Layer Perceptron module (MLP). For the self-attention module in encoder, query, and key elements are pixels in the feature maps with positional encodings.

Similar to an encoder, a Transformer decoder also consists of $K$ decoder layers. Each decoder layer comprises a multi-head self-attention module, a multi-heads cross-attention module, and two MLP modules. Except for the state, $z$, generated by the encoder, the Transformer decoder takes a fixed number of so-called positional embeddings as input. Then, the decoder generates a fixed number of output embeddings. Also, the decoder is permutation-invariant. Thus, the positional embeddings must be different from each other, so that the decoder produces various output embeddings. Finally, the output embedding is decoded into a fixed number of points, which consists of the class, center coordinates, and the radius through the prediction heads.

\subsection{Set-Based Loss}
Through a prediction head, NRTR generates a fixed-size set of $N$ points to represent the morphological characteristics of neurons in the original image block. 

To simplify the discussion regarding set-based loss, we define a new class, "no-object" ($\varnothing$). In the training stage, "no-object" points are involved in the loss function introduced below. In the inference stage, "no-object" points are directly removed. Any fixed-size prediction set, $\hat{y}$, can be viewed as a ground truth set, $y$, which has been padded with "no-object" points if the size of $\hat{y}$ is not larger than the size of $y$.

As shown in Fig. \ref{Fig-4}, predicted classes, center coordinates, and radiuses of $N$ points are compared with those of $M$ ground truth points. Before comparison, the best-matched prediction point is assigned to each ground truth point. The overall process is formulated as a minimum weight matching problem of a bipartite graph. Thus, if the ground truth set contains only $M$ points, the Hungarian algorithm is used to choose $M$ best-matched points involved in the loss; while the other $N-M$ points are discarded.

Assuming $N$ is larger than the number of points in any ground truth set, we consider $y$ as a set of size $N$ padded with "no-object". Each point consists of center coordinates, radius, and class; therefore, we define loss between a prediction point and a ground truth point, $L_{point}(y_i,\hat{y}_i)$, as follows:
\begin{equation}
\begin{aligned}
L_{point}(y_i,\hat{y}_i) = -\mathbbm{1}_{\{cls_i \ne \varnothing \}}w_{cls}L_{cls}(cls_i, \hat{cls}_i)\\
+ \mathbbm{1}_{\{cls_i \ne \varnothing \}}w_{box} L_{box}(y_i, \hat{y}_i)\\
+ \mathbbm{1}_{\{cls_i \ne \varnothing \}}w_{iou} L_{iou}(y_i, \hat{y}_i)\\
\end{aligned}
\end{equation}
where $L_{cls}(cls_i, \hat{cls}_i) $ is the classification loss; whereas, $L_{box}(y_i, \hat{y}_i)$ and $L_{iou}(y_i, \hat{y}_i)$ are the regression loss between the $i^{th}$ ground truth and $i^{th}$ prediction, respectively. Hyperparameters are $w_{cls}, w_{box}, w_{iou}$; $L_{iou}(\cdot,\cdot)$ is defined as the generalized IoU loss proposed by Rezatofighi \textit{et al.} \cite{rezatofighi2019generalized}; $L_{box}(\cdot,\cdot)$ is the most commonly-used L1 loss.

To find the minimum weight bipartite match between these two sets, we search for a permutation of $N$ points, $\sigma$. The optimal assignment, $\hat{\sigma}$, is calculated with the Hungarian algorithm.
\begin{equation}
\hat{\sigma}=\mathop{\arg\min}_{\sigma}\sum_i^M L_{point}(y_i,\hat{y}_{\sigma(i)})\label{eq5}
\end{equation}

On the basis of the minimum weight bipartite match, $\sigma$, and point loss, $L_{point}(y_i,\hat{y}_{\sigma(i)})$, we define the set-based loss between prediction set and ground truth set, $ L_{set}(y,\hat{y})$, as follows:
\begin{equation}
L_{set}(y,\hat{y}) = \sum_i^M L_{point}(y_i,\hat{y}_{\hat{\sigma}(i)})
\end{equation}
where $L_{point}(y_i,\hat{y}_{\sigma(i)})$ is a matching cost between the point of ground truth set, $y_i$, and the corresponding point of prediction set, $\hat{y}_{\sigma(i)}$.

\section{Experiment and Result}
To verify the effectiveness of our proposed NRTR, we conducted extensive experiments with various datasets. Results are shown our results for the VISoR-40 dataset \cite{zhao2019progressive} and the BigNeuron dataset \cite{2015BigNeuron}. The neuron reconstruction results of NRTR are compared with those of four neuron segmentation models on these two datasets. NRTR achieves a better Precision-Recall trade-off on overlap metrics. Additionally, a scale study is provided to demonstrate the effectiveness of our method on these datasets. Our proposed NRTR, adapted from DETR, has much room for further improvement.

\subsection{Datasets}
\subsubsection{VISoR-40 Dataset}
The VISoR-40 dataset \cite{zhao2019progressive} consists of forty image blocks cropped from an optical microscopy image captured from a mouse brain. The physical resolution of the whole image stack is $0.5\times0.5\times0.5 um^3$ per voxel. The size of the image stacks ranges from $419 \times 1197 \times 224 $ to $869 \times 1853 \times 575$. All image stacks of the dataset are captured by the VISoR imaging system. The raw image stacks are 16-bit; therefore, enough neurite details are preserved for neuron morphology reconstruction. There are 8 image stacks annotated manually by neuroscience experts and 32 raw image stacks. In our experiments, the annotated dataset was partitioned into six training and two test image stacks.

\subsubsection{BigNeuron Dataset}
The BigNeuron project \cite{2015BigNeuron} is sponsored by fourteen neuroscience-related research organizations and dozens of international research groups and individuals. The BigNeuron dataset contains 166 neurons with standard reconstruction results and respective raw image stacks. These image datasets, originally contributed by a number of labs around the world, were standardized and reconstructed by six or seven annotators during the BigNeuron Annotation workshop held at Allen Institute for Brain Science, Seattle, June 2015. The dataset comprises neuron image stacks and their respective annotations from different species and nervous system regions. Some neurons come from famous neuroinformatics projects, such as the Allen Mouse and Human Cell Type Project, Taiwan FlyCircuits, and Janelia Fly Light. Others are contributed directly by neuroscientists worldwide. The BigNeuron dataset consists of only image stacks of single-neuron or neurons that have clear separation. Because neuron image stacks can be 8-bit or 16-bit, we upsampled all images to 16-bit to be consistent with experiments using the VISoR-40 dataset. To evaluate the reconstruction results, we randomly partitioned the annotated dataset into training and test sets.

\subsection{Implementation Details}
Experiments were conducted using the VISoR-40 dataset and BigNeuron dataset. Due to the size of the whole OM image, it was impractical to train and infer with the whole image. In the training and inference stages, fixed-size patches were cropped continuously from the original images to fit the input size of the NRTR. In experiments with both the VISoR-40 dataset and the BigNeuron dataset, the input patch size of NRTR is $64 \times 64 \times 64$ voxels.

Our model was trained for 100 epochs by setting the initial learning rates of the Transformer and backbone to $10^{-4}$ and $10^{-5}$, respectively, and weight decay to $10^{-4}$. An Adam optimizer \cite{Diederik2015adam} was employed for 100 epochs with a cosine decay learning rate scheduler and linear warmup of 10 epochs.

The backbone is with a different-scale ResNet model \cite{he2016deep}, varying from ResNet18 to ResNet50. For higher training efficiency, we discarded empty image blocks (background) during the training stage. Besides, we apply data augmentation using rotation and flipping. NRTR is trained and evaluated on a computer with two NVIDIA 3090 GPUs.

\subsection{Evaluation Metrics}
To evaluate the performance of our proposed NRTR, we converted neuron reconstructions to neuron segmentation results through the standard Vaa3d plugin \cite{peng2010v3d}. We adopted four indicators for quantitative analysis: Precision, Recall, F-Score, and Jaccard. 

\subsection{Experiments on the VISoR-40 Dataset}
Only eight annotated image stacks are used for model training and testing. In the experiments with the VISoR-40 dataset, we trained the NRTR with six image stacks and evaluated the other image stacks.

To compare the performance of NRTR with other neuron segmentation methods, the neuron reconstruction results generated by NRTR are converted to segmentation results through the standard Vaa3d plugin. We compared different-scale NRTR against the following previous state-of-the-art on overlap criteria: SGSNet \cite{yang2021structure}, MP-NRGAN \cite{chen2021weakly}. Neuron reconstruction results with two different backbones:(ResNet34 and ResNet50) were reported. The corresponding models are NRTR-ResNet34 and NRTR-ResNet50, respectively. Table \ref{tab:table1} presents the comparison with other neuron segmentation methods on the VISoR-40 dataset. 

\begin{table}[ht]
\caption{Comparison with state-of-the-art deep learning methods on the BigNeuron dataset. We build overlap criteria between the prediction and the target.}
    \centering
    \begin{tabular}{c c c c c}
        \toprule
        Method & Precision & Recall & F-Score & Jaccard \\
        \midrule
        Li-2017 \cite{li2017deep}        & 0.4158 & 0.5057 & 0.4563 & 0.2956 \\
        Huang-2020 \cite{huang2020weakly}& 0.4261 & 0.4326 & 0.4293 & 0.2733 \\
        MP-NRGAN \cite{chen2021weakly}    & 0.4283 & 0.472  & 0.4491 & 0.2893 \\
        SGSNET \cite{yang2021structure}  & 0.475  & 0.365  & 0.4128 & 0.2601 \\
        \midrule
        NRTR-ResNet34 &  0.4633  &  0.5526  &  0.5034  &  0.3363  \\
        NRTR-ResNet50 & \textbf{0.4755}  &  \textbf{0.5556}  &  \textbf{0.5124}  &  \textbf{0.3445} \\
        \bottomrule
    \end{tabular}
\label{tab:table1}
\end{table}

\begin{figure*}[ht]
\centerline{\includegraphics[width=\textwidth]{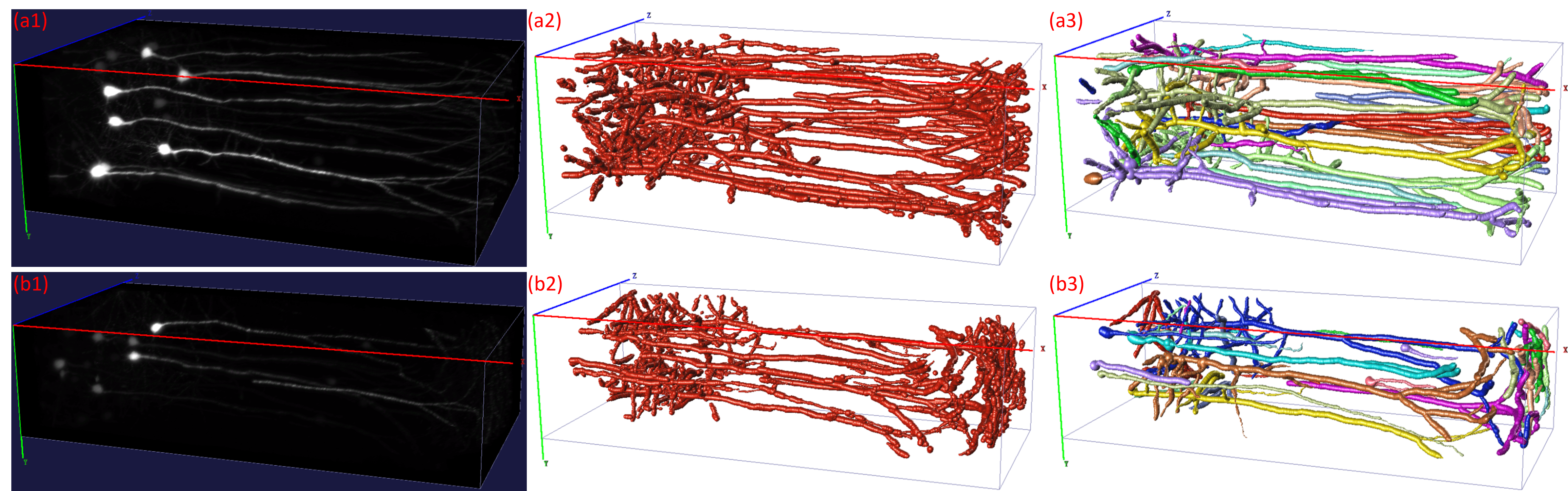}}
\caption{Neuron reconstruction results of the NRTR on two test images from VISoR-40 Dataset. (a1) (b1) raw neuron image stacks. (a2) (b2) 3D models produced by neuron reconstruction results of NRTR. (a3) (b3) 3D models produced by ground truth.}
\label{Fig-5}
\end{figure*}

Compared with previous state-of-to-art methods, NRTR-ResNet34 surpasses MP-NRGAN: +3.5\% on Precision, +8.06\% on Recall, +5.43\% on F-Score, and +4.7\% on Jaccard. NRTR-ResNet50 surpasses MP-NRGAN: +4.72\% on Precision, +8.36\% on Recall, +6.33\% on F-Score, and +5.52\% on Jaccard.

Fig. \ref{Fig-5} shows that the topological structure of neuron reconstruction results is similar to that of the ground truth, except for some subtle noises, indicating that NRTR extracts topological information of interest from the raw image stacks, although NRTR differs considerably from previous neuronal reconstruction methods, which detect foreground voxels.

\subsection{Experiments on the BigNeuron Dataset}
Compared with the VISoR-40 dataset, the BigNeuron dataset (provided by Allen Institute) is much larger-scale. In the experiments using the BigNeuron dataset, to evaluate the performance of our model, the dataset was partitioned randomly into training and test sets. Due to the low-resolution image stacks in the BigNeuron dataset, which greatly influence model performance, linear interpolation was applied to magnify the original neuron image stack by a factor of eight.
\begin{table}[ht]
\caption{Comparison with state-of-the-art deep learning methods on the VISoR-40 dataset. We build overlap criteria between neuron reconstruction result and manual annotation.}
    \centering
    \begin{tabular}{c c c c c}
        \toprule
        Method & Precision & Recall & F-Score & Jaccard \\
        \midrule
        Li-2017 \cite{li2017deep}          & 0.7073 & 0.5197 & 0.5979  & 0.4264 \\
        Huang-2020 \cite{huang2020weakly} & 0.5964 & 0.508  & 0.5487  & 0.378 \\
        MP-NRGAN \cite{chen2021weakly}  & 0.7117 & 0.5563 & 0.6245  & 0.454 \\
        SGSNET \cite{yang2021structure}   & 0.692  & 0.5047 & 0.5837  & 0.4121 \\
        \midrule
        NRTR-ResNet34 & 0.7044 &  \textbf{0.6533}  &  \textbf{0.6779}  &  \textbf{0.5127} \\
        NRTR-ResNet50 & \textbf{0.7173} & 0.6164 & 0.6630 & 0.4959 \\
        \bottomrule
    \end{tabular}
\label{tab:table2}
\end{table}

\begin{figure*}[ht]
\centerline{\includegraphics[width=\textwidth]{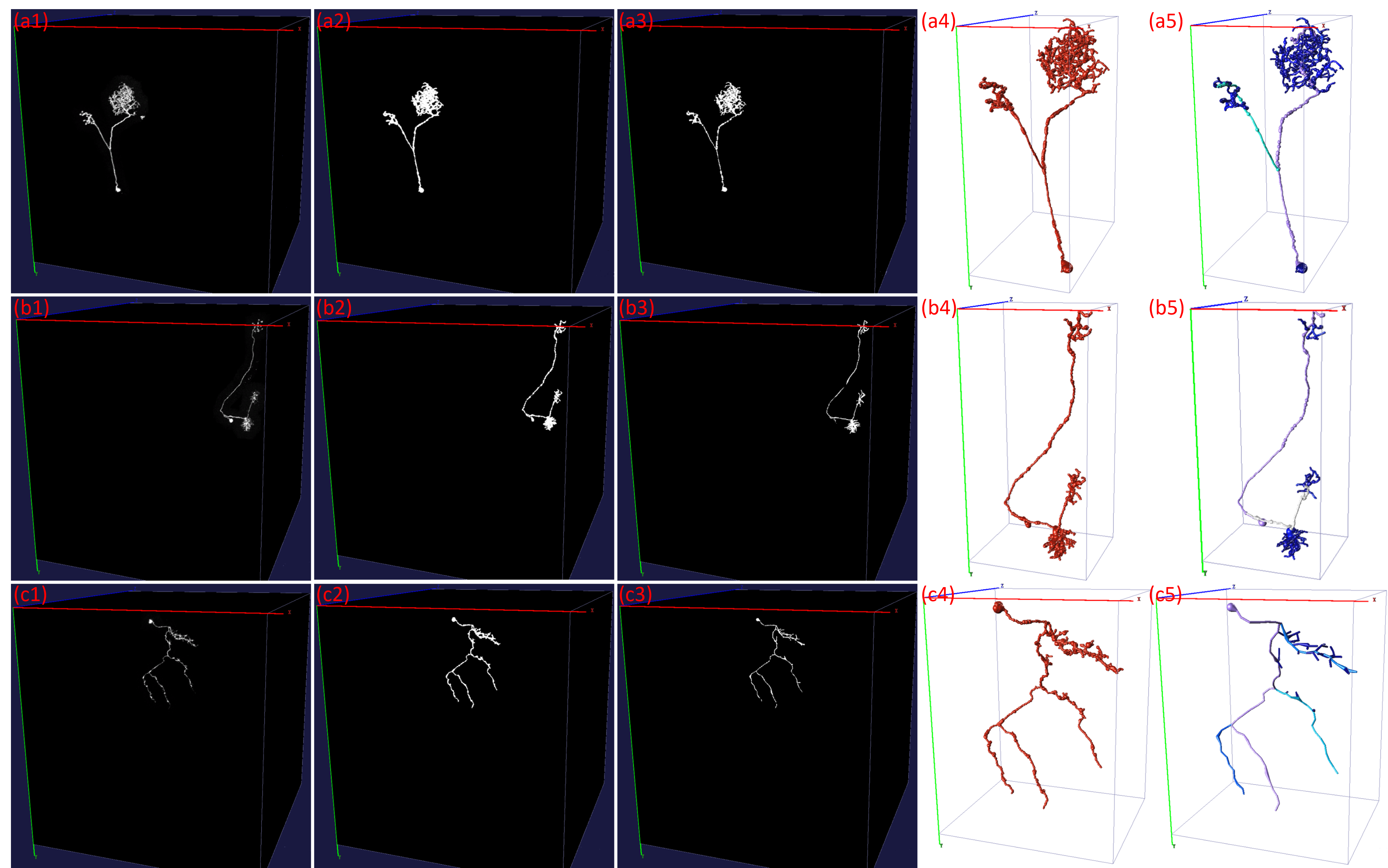}}
\caption{Neuron reconstruction results of the NRTR on two test images from BigNeuron Dataset. (a1) (b1) (c1) Raw neuron image stacks. (a2) (b2) (c2) Masks produced by neuron reconstruction results of NRTR. (a3) (b3) (c3) Masks produced by ground truth. (a4) (b4) (c4) 3D models produced by neuron reconstruction results of NRTR. (a5) (b5) (c5) 3D models produced by ground truth.}
\label{Fig-6}
\end{figure*}

As with the experiments using the VISoR-40 dataset, we compared NRTR with four deep learning methods. Neuron reconstruction results are reported with two different backbones: ResNet34 and ResNet50.

Table \ref{tab:table2} shows that NRTR achieves state-of-the-art performance on the BigNeuron dataset. NRTR-ResNet34 surpasses MP-NRGAN: -0.74\% on Precision, +9.7\% on Recall, +5.34\% on F-Score, and +5.84\% on Jaccard. NRTR-ResNet50 surpasses MP-NRGAN: +0.56\% on Precision, +6.01\% on Recall, 3.85\% on F-Score, and +4.19\% on Jaccard.

Fig. \ref{Fig-6} shows the comparison among the 3D models and masks produced by neuron reconstruction results of NRTR and the 3D models and masks produced by ground truth. Unlike the VISoR-40 dataset, each image stack in the BigNeuron dataset contains only a single neuron or disconnected multiple neurons, separated with relative clarity. Therefore, the morphological characteristics of neuron image stacks in the VISoR-40 dataset are more complex than in the BigNeuron dataset. By comparing the 3D models of the reconstruction results on the two datasets, we found that NRTR can generate more high-quality reconstructions on the BigNeuron dataset rather than the VISoR-40 dataset. However, in terms of various metrics, it appears that NRTR performs better on the VISoR-40 dataset (See in Table \ref{tab:table1} and Table \ref{tab:table2}). We attribute this phenomenon to the different resolutions of the original images in the two datasets. A low-resolution image indicates that the radiuses of most points in the annotation are too small. A tiny deviation in 3D center coordinates and radiuses of predicted points leads to a significant decrease in Precision and an increase in Recall. In addition, because the backbone transforms the original low-resolution image block into a lower-resolution feature map, the spatial information of the region corresponding to minor points is likely to be lost.

\subsection{Discussion}
Thanks to the global image information achieved by the self-attention mechanism, NRTR has significant advantages in high-resolution neuron image stacks over the state-of-the-art approaches mentioned above. Thus, we use a linear interpolation algorithm on low-resolution image stacks to increase image resolution before model inference.

Extensive experiments reveal that, when the input image block size is $64\times64\times64$, and the radius of most points is more than six, NRTR achieves desirable reconstruction results. In other words, the ratio of the point radius to the input image block size should not be too small. This novel image-to-set model design for neuron reconstruction also comes with new challenges, particularly regarding training and reconstruction results on low-resolution image blocks.

In NRTR, we apply a rule-based algorithm to construct connection relationships among the points of a set. Our aim is to design a deep learning module for learning connection information. We anticipate future work will address this successfully.

\section{Conclusion}
We proposed a novel design for neuron reconstruction, NRTR. We employed an image-to-set model to accomplish the neuron reconstruction automatically, compared with the previous U-Net-based model. Transformer architecture makes end-to-end neuron reconstruction available, thereby greatly reducing pipeline complexity and making model training more convenient. On one hand, our proposed approach achieves excellent neuron reconstruction results with 3D OM neuronal images. Our model achieves results comparable to the state-of-the-art approaches with VISoR-40 and BigNeuron datasets (See Table \ref{tab:table1} and Table \ref{tab:table2}).

On the other hand, NRTR performs less satisfactorily on low-resolution image stacks. We try to utilize an image interpolation algorithm to mitigate the issue. Besides, our proposed method applies connectivity construction to construct connection relationships. It might be better to design a deep learning module to replace it. We hope future work will lead to more effective solutions.

\bibliographystyle{ieeetr}
\bibliography{main}
\end{document}